\documentclass[10pt,twocolumn,letterpaper]{article}
\usepackage{iccv}
\usepackage{times}
\usepackage{epsfig}
\usepackage{graphicx}
\usepackage{amsmath}
\usepackage{nicefrac}
\usepackage{amssymb}
\usepackage{enumitem}
\usepackage[normalem]{ulem}
\usepackage[LGRgreek]{mathastext}
\usepackage{afterpage}

\usepackage[breaklinks=true,bookmarks=false]{hyperref}
\hypersetup{
    colorlinks=true,
    linkcolor=blue,
    filecolor=magenta,      
    urlcolor=blue,
}

\iccvfinalcopy 

\def\iccvPaperID{22} 

\ificcvfinal\fi
\begin{document}

\title{RITnet: Real-time Semantic Segmentation of the Eye for Gaze Tracking}

\author{ Aayush K.Chaudhary\thanks{Equal Contribution.} \quad Rakshit Kothari$^{*}$ \quad Manoj Acharya$^{*}$ \quad Shusil Dangi \quad \\ Nitinraj Nair \quad Reynold Bailey \quad Christopher Kanan \quad Gabriel Diaz \quad Jeff B. Pelz\\
Rochester Institute of Technology, USA\\
{\tt\small(akc5959,rsk3900,ma7583,sxd7257,nrn2741,rjbvcs,kanan,gabriel.diaz,jeff.pelz)@rit.edu}}

\maketitle
\ificcvfinal\thispagestyle{empty}\fi


\begin{abstract}
Accurate eye segmentation can improve eye-gaze estimation and support interactive computing based on visual attention; however, existing eye segmentation methods suffer from issues such as person-dependent accuracy, lack of robustness, and an inability to be run in real-time.
Here, we present the RITnet model, which is a deep neural network that combines U-Net and DenseNet. RITnet is under 1 MB and achieves 95.3\% accuracy on the 2019 OpenEDS Semantic Segmentation challenge. Using a GeForce GTX 1080 Ti, RITnet tracks at $>$ 300Hz, enabling real-time gaze tracking applications. Pre-trained models and source code are available \footnote{\url{https://bitbucket.org/eye-ush/ritnet/}}.
\end{abstract}

\section{Introduction}
Robust, accurate, and efficient gaze estimation is required to support a number of critical applications such as foveated rendering, human-machine and human-environment interactions, as well as inter-saccadic manipulations, such as redirected walking~\cite{Sun2018TowardsWalking}. Recent non-intrusive, video-based eye-tracking methods involve localization of eye features such as the pupil~\cite{Kassner2014Pupil:Interaction} and/or iris~\cite{Wood2014EyeTab:Computers}. These features are then regressed onto some meaningful representation of an individual's gaze. Convolutional neural networks (CNNs) have demonstrated high accuracy ~\cite{Kim2019NVGaze:Estimation, Wu2019EyeNet:Understanding} and robustness in unconstrained lighting conditions ~\cite{B2019500000Segmentation} and an ability to generalize under low resolution constraints~\cite{Park2019Few-shotEstimation, Park2018DeepEstimation}.

\begin{figure}[t]
\begin{center}
\includegraphics[width=8.3cm]{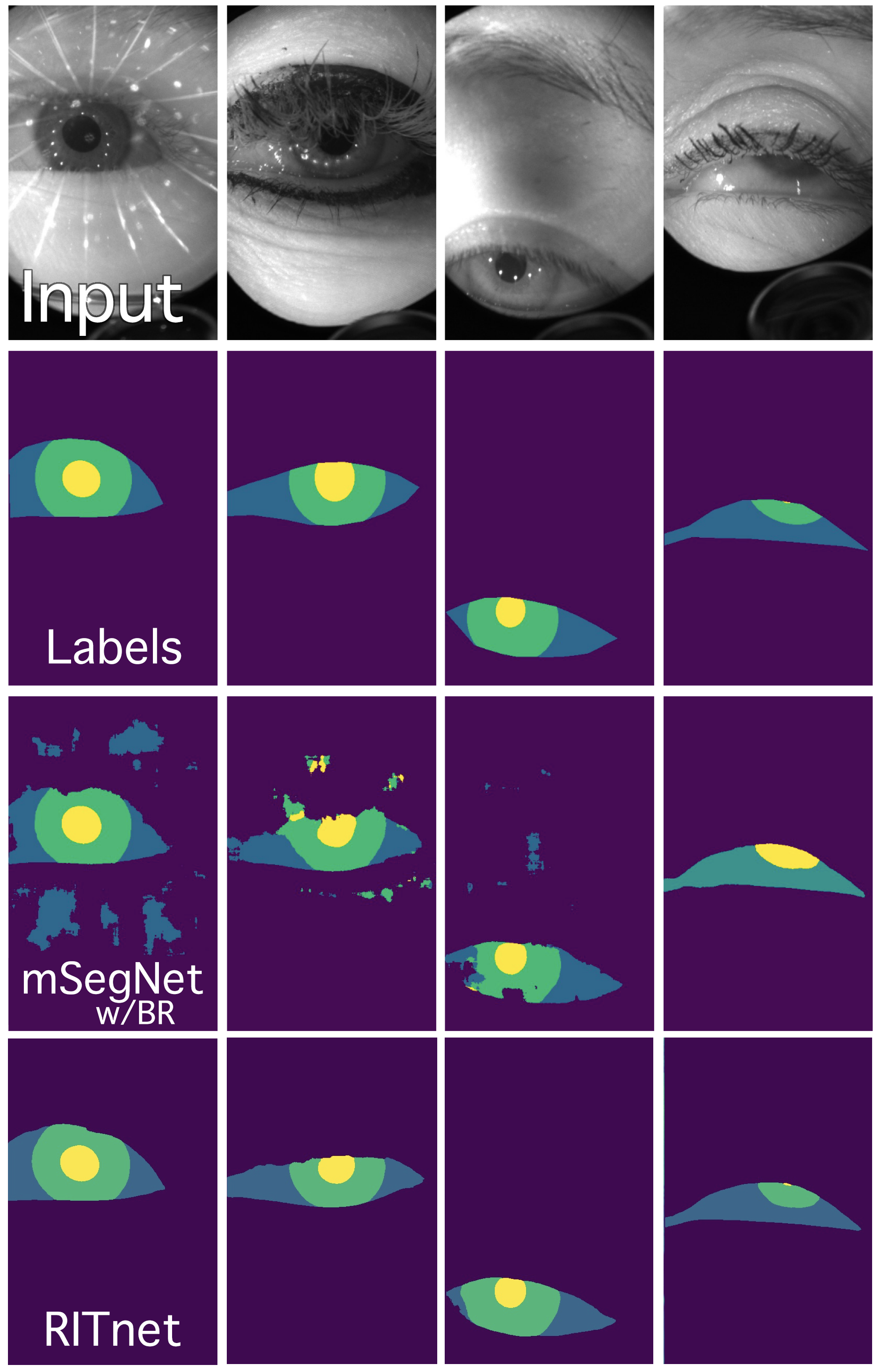}
\end{center}
   \caption{Comparison of model performance on difficult samples in the OpenEDS test-set. Top-row left to right shows eyes obstructed due to prescription glasses, heavy mascara, dim light and partial eyelid closure. Rows from top to bottom show input test images, ground truth labels, predictions from mSegNet w/BR~\cite{Garbin2019OpenEDS:Dataset} and predictions from RITnet, respectively.}
   
\label{fig:output}
\end{figure}

In an effort to engage the machine learning and eye-tracking communities in the field of eye-tracking for head-mounted displays (HMD), Facebook Reality Labs issued the Open Eye Dataset (OpenEDS) Semantic Segmentation challenge which addresses part of the gaze estimation pipeline: 
identifying different regions of interest (e.g., pupil, iris, sclera, skin) in close-up images of the eye. 
Such \textit{semantic segmentation} of these regions enables the extraction of region-specific features (e.g., iridial feature tracking~\cite{Chaudhary2019MotionMovements})and mathematical models which summarize the region structures (e.g., iris ellipse~\cite{Wood2014EyeTab:Computers, B2019500000Segmentation, Park2018DeepEstimation}, or pupil ellipse~\cite{Kassner2014Pupil:Interaction}) used to derive a measure of gaze orientation.

\paragraph{The major contributions of this paper are as follows:}
\begin{enumerate}[noitemsep,nolistsep]
    \item We present RITnet, a semantic segmentation architecture that obtains state-of-the-art results on the 2019 OpenEDS Semantic Segmentation Challenge with model size of \textbf{only 0.98 MB}. Our model performs segmentation at 301 $Hz$ for 640x400 images on an NVIDIA 1080Ti GPU.
    \item We propose domain-specific augmentation schemes which help in generalization under a variety of challenging conditions.
    \item We present boundary aware loss functions with a loss scheduling strategy to train Deep Semantic Segmentation models. This helps in producing coherent regions with crisp region boundaries.
    
\end{enumerate}

\begin{figure}[t]
\begin{center}
\includegraphics[width=8.3cm]{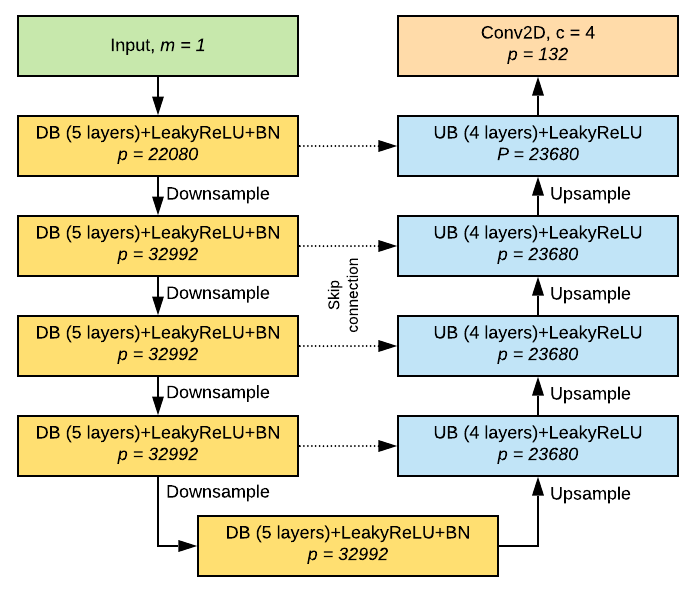}
\end{center}
\caption{Architecture details of RITnet. DB refers to \textit{Down}-Block, UB refers to \textit{Up}-Block, and BN stands for batch normalization. Similarly, \textit{m} refers to the number of input channels ($m=1$ for gray scale image), \textit{c} refers to number of output labels and \textit{p} refers to number of model parameters. Dashed lines denote the skip connections from the corresponding \textit{Down}-Blocks. All of the Blocks output tensors of channel size  $m$=32.}
\label{fig:parameters}
\end{figure}

\section{Previous Works}

Recently developed solutions for end-to-end segmentation involve using Deep CNNs to produce a labeled output irrespective of the size of the input image. Such architectures consist of convolution layers with a series of down-sampling followed by progressive upsampling layers. Downsampling operations strip away finer information that is crucial for accurate pixel-level semantic masks. This limitation was mitigated by Ronneberger et al. by introducing skip-connections between the encoder and decoder~\cite{Ronneberger2015U-net:Segmentation}. Jergou et al. proposed TiramisuNet~\cite{Jegou2017TheSegmentation}, a progression of dense blocks~\cite{Huang2017DenselyNetworks} with skip connections between the up- and down-sampling pathways. TiramisuNet demonstrated reuse of previously computed feature maps to minimize the required number of parameters.  Dangi et al. proposed the DenseUNet-K architecture~\cite{Dangi2019} for image-to-image translation based on simplified dense connected feature maps with skip connections. The RITnet model presented in this paper is based on the DenseUNet-K architecture\footnote{\url{https://github.com/ShusilDangi/DenseUNet-K}}.

\section{Proposed Model: RITnet}
Recently, segmentation models based on Fully Convolutional Networks (FCN) have performed well across many datasets \cite{Jegou2017TheSegmentation, Ronneberger2015U-net:Segmentation}. That success, however, often comes at the cost of computational complexity, restricting their feasibility for real-time applications where rapid computation and robustness to illumination conditions is paramount~\cite{Garbin2019OpenEDS:Dataset}. In contrast, RITnet has 248,900 trainable parameters which require less than 1MB storage with 32-bit precision (see Figure~\ref{fig:parameters}) and has been benchmarked at $>$300 $Hz$.

RITnet has five \textit{Down}-Blocks and four \textit{Up}-Blocks which downsample and upsample the input. The last \textit{Down}-Block is also referred to as the \textit{bottleneck} layer which reduces the overall information into a small tensor $\nicefrac{1}{16}^{th}$ of the input resolution. Each \textit{Down}-Block consists of five convolution layers with LeakyReLU activation. All convolution layers share connections from previous layers inspired by DenseNet~\cite{Huang2017DenselyNetworks}. We maintain a constant channel size as in DenseUNet-K \cite{Dangi2019} with K=32 channels to reduce the number of parameters.  All \textit{Down}-Blocks are followed by an average pooling layer of size 2x2. The \textit{Up}-Block layer upsamples its input by a factor of two using the nearest neighbor approach. Each \textit{Up}-Block consists of four convolution layers with LeakyReLU activation. All \textit{Up}-Blocks receive extra information from their corresponding \textit{Down}-Block via skip connections, an effective strategy which provides the model with representations of varying spatial granularity. 

\subsection{Loss functions}
Each pixel is classified into one of four semantic categories: \textit{background, iris, sclera,} or \textit{pupil}. Standard cross-entropy loss (CEL) is the default choice for applications with a balanced class distribution. However, there exists an imbalanced distribution of classes with the fewest pixels representing pupil regions. While CEL aims to maximize the output probability at a pixel location, it remains agnostic to the structure inherent to eye images. To mitigate these issues, we implemented the following loss functions:

\textbf{Generalized Dice Loss (GDL):} Dice score coefficient measures the overlap between the ground truth pixel and their predicted values. In cases of class imbalance~\cite{Milletari2016V-Net:Segmentation}, weighting the dice score by the squared inverse of class frequency~\cite{Sudre2017GeneralisedSegmentations} showed increased performance when combined with CEL.

\textbf{Boundary Aware Loss (BAL)}: 
Semantic boundaries separate regions based on class labels.  Weighting the loss for each pixel by its distance to the two nearest segments introduces edge awareness~\cite{Ronneberger2015U-net:Segmentation}. We generate boundary pixels using a Canny edge detector which are further dilated by two pixels to minimize confusion at the boundary. We use these edges to mask the CEL.

\textbf{Surface Loss (SL)}: 
 SL is based on a distance metric in the space of image contours which preserves small, infrequent structures of high semantic value~\cite{Kervadec2018BoundarySegmentation}. BAL attempts to maximize the correct pixel probabilities near boundaries while GDL provides stable gradients for imbalanced conditions. Contrary to both, SL scales the loss at each pixel based on its distance from the ground truth boundary for each class. It is effective in recovering smaller regions which are ignored by region based losses~\cite{Kervadec2018BoundarySegmentation}. 

The total loss $\mathcal{L}$ is given by a weighted combination of these losses as $
\mathcal{L} = 
\mathcal{L}_{CEL} (\lambda_1 + \lambda_2 \mathcal{L}_{BAL} ) + \lambda_3 \mathcal{L}_{GDL} + \lambda_4 \mathcal{L}_{SL}$.

\section{Experimental Details}
\subsection{Dataset and Evaluation}
We train and evaluate our model on the OpenEDS Semantic Segmentation dataset~\cite{Garbin2019OpenEDS:Dataset} consisting of 12,759 images split into \textit{train} (8,916), \textit{validation} (2,403) and \textit{test} (1,440) subsets. Each image had been hand annotated with four semantic labels; \textit{background, sclera, pupil, \& iris}.

Per OpenEDS challenge guidelines, our \textit{overall score} metric uses the average of the mean Intersection over Union (mIoU) metric for all classes and model size (S) calculated as a function of number of trainable parameters in megabytes (MB). The \textit{overall score} is given as 
$\frac{mIoU + \textrm{min}(\nicefrac{1}{S}, 1) }{2}$.

\subsection{Training}
We trained our model using Adam~\cite{Kingma2014Adam:Optimization} with a learning rate of 0.001 and a batch size of 8 images for 175 epochs on a TITAN 1080 Ti GPU. We reduced the learning rate by a factor of 10 when the validation loss plateaued for more than 5 epochs. The selected model with the best validation score was found at the $151^{st}$ epoch. In our experiments, we used $\lambda_1=1,\lambda_2=20,\lambda_3=(1-\alpha)$ and $\lambda_4=\alpha$,
where $\alpha = epoch / 125$ for epoch$<$125 otherwise 0. This loss scheduling scheme gives prominence to GDL during initial iterations until a steady state is achieved, following which SL begins penalizing stray patches. 

\subsection{Data Pre-processing}

\begin{figure}[h]
\begin{center}
\includegraphics[width=8.3cm]{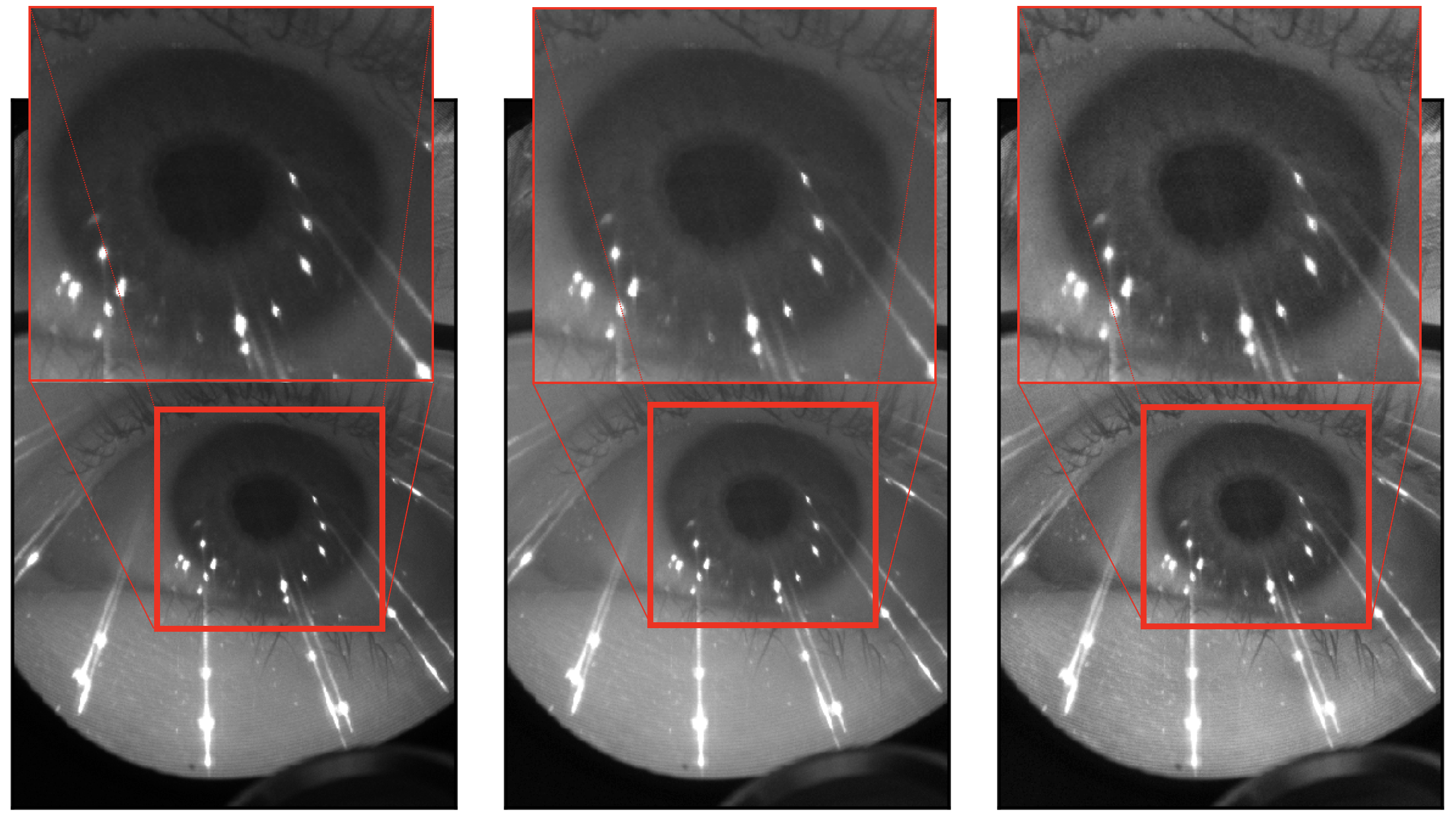}
\end{center}
   \caption{Left to right: Original image, image after gamma correction, image after CLAHE is applied. Note that in the rightmost image, it is comparatively easier to distinguish iris and pupil.}
\label{fig:preprocess}
\end{figure}

To accommodate variation in individual reflectance properties (e.g., iris pigmentation, eye makeup, skin tone or eyelids/eyelashes)~\cite{Garbin2019OpenEDS:Dataset} and HMD specific illumination (the position of infrared LEDs with respect to the eye), we performed two pre-processing steps. These steps were based on the difference in the train, validation and test distributions of mean image brightness (Figure 11 in Garbin et. al~\cite{Garbin2019OpenEDS:Dataset}).Pre-processing reduced these differences and also
increased separability of certain eye features. First, a fixed gamma correction with an exponent of 0.8 was applied to all input images. Second, we applied local Contrast Limited Adaptive Histogram Equalization (CLAHE) with a grid size of 8x8 and clip limit value of 1.5~\cite{Zuiderveld1994ContrastEqualization}. Figure~\ref{fig:preprocess} shows an image before and after pre-processing.

\begin{figure}[h]
\begin{center}
\includegraphics[width=8.3cm]{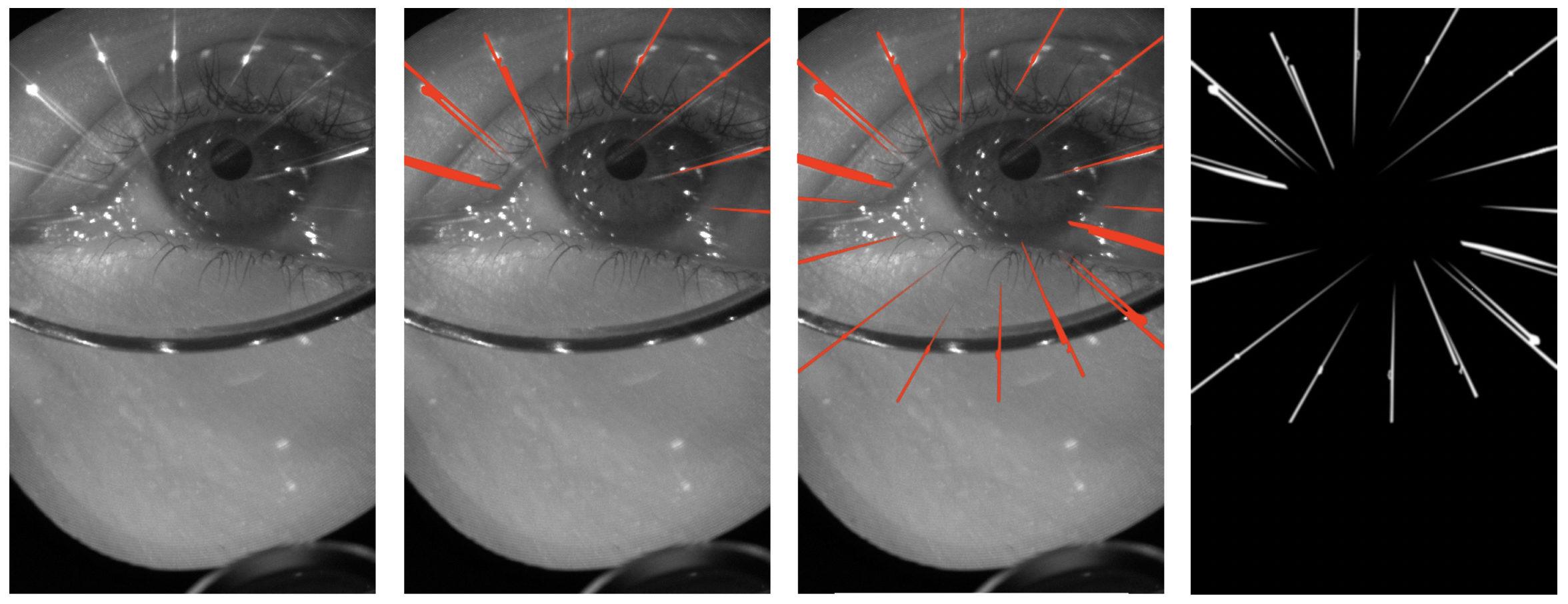}
\end{center}
   \caption{Generation of a \textit{starburst} pattern from the training image 000000240768. Left to Right: Original image, selected reflections, concatenating with its 180$^{\circ}$ rotation, final pattern mask (best viewed in color).}
\label{fig:starburst}
\end{figure}

To increase the robustness of the model to variations in image properties, training data was augmented with the following modifications:
\begin{itemize}
\item Reflection about the vertical axis.
\item Gaussian blur with a fixed kernel size of 7x7 and standard deviation $2\leq\sigma\leq7$. 
\item Image translation of 0-20 pixels in both axes. 

\item Image corruption using 2-9 thin lines drawn around a random center  ($120<x<280,192<y<448$)

\item Image corruption with a structured \textit{starburst} pattern (Figure~\ref{fig:starburst}) to reduce segmentation errors caused by reflections from the IR illuminators on eyeglasses. Note that the \textit{starburst} image is translated by 0-40 pixels in both directions.

\end{itemize}

Each image received at least one of the above-mentioned augmentations with a probability of 0.2 on each iteration. The probability that an image would be flipped horizontally was 0.5.

\section{Results}
\begin{figure}[h]
\begin{center}
\includegraphics[width=8.3cm]{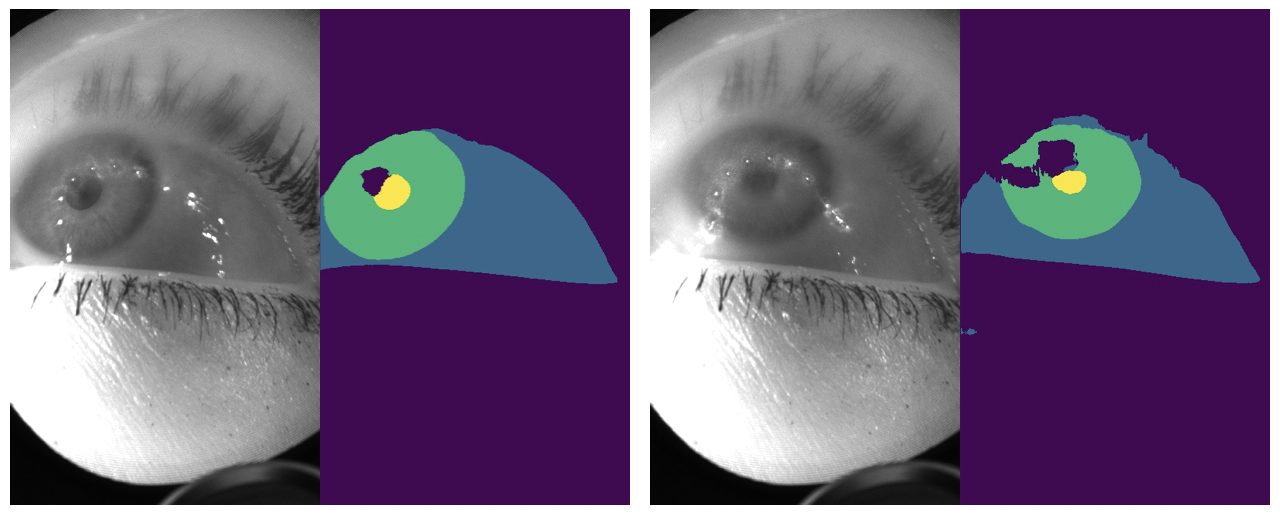}
\end{center}
   \caption{ Our model struggles to do an accurate segmentation when eye masks are heavily blurred or defocused.}
\label{fig:output2}
\end{figure}

We compare our results against SegNet~\cite{Garbin2019OpenEDS:Dataset}, another fully convolutional encoder-decoder architecture. mSegNet refers to the modified SegNet with four layers of encoder and decoder. mSegNet w/BR refers to mSegNet with Boundary Refinement as residual structure and mSegNet w/SC is a lightweight mSegNet with depthwise Separable Convolutions ~\cite{Garbin2019OpenEDS:Dataset}. As shown in Table~\ref{tbl:result}, our model achieves a $\sim$6\% improvement in mIoU score while the complexity is reduced by $\sim$38\% compared to the baseline model mSegNet w/SC. However, our model's segmentation quality was impacted at higher values of motion blur and image defocus (Figure~\ref{fig:output2}), Figure~\ref{fig:output} demonstrates that our model generalizes to some challenging cases where other models fail to produce coherent results.

\begin{table}[h]
\begin{center}
\resizebox{\linewidth}{!}
{%
\begin{tabular}{|c|c|c|c|c|c|}
\hline
Model  & Mean & mIoU & Model    & No. of & Overall\\
       & F1   &  & Size & parameters & Score\\
              &  &  & (S)  & (million) &\\
\hline
mSegNet* & 97.9 & 90.7 & 13.3 & 3.5 & 0.491\\
\hline
mSegNet* & 98.3 & 91.4 & 13.3 & 3.5 & 0.495\\
w/BR  &  &  &  &  & \\
\hline
mSegNet*  & 97.4 & 89.5 & 1.6 & 0.4 & 0.762\\
w/SC(B)  &  &  &  &  & \\
\hline
\textbf{Ours} & \textbf{99.3} & \textbf{95.3} & \textbf{0.98} & \textbf{0.25} & \textbf{0.976}\\
\hline
\end{tabular}
}%
\end{center}
\caption{Performance comparison on the test split of the OpenEDS dataset. The metrics and comparison models (*) are used as reported in ~\cite{Garbin2019OpenEDS:Dataset}.}
\label{tbl:result}
\end{table}

\section{Discussion} 
Our model achieves state-of-the-art performance with a small model footprint.The final architecture was arrived at after exploring a number of architectural variations. Reducing the channel size from 32 to 24 and increasing the number of convolution layers in the \textit{Down}-Block did not affect the results. Surprisingly, increasing the channel size to 40 and removing one convolutional layer in the \textit{Down}-Block degraded performance, resulting in spurious patches in output regions. Performance was influenced by the choice of loss functions and the adjustment of their relative weights. By setting the boundary-aware loss at a relatively higher weight, we observed sharp boundary edges and consequently improved our test mIoU from 94.8\% to 95.3\%.

We speculate that some aspects of our model were successful because they accounted for labeling artifacts in the openEDS dataset. For example, although pupil-to-iris boundaries were defined using ellipse fits to multiple points selected on the boundaries~\cite{Garbin2019OpenEDS:Dataset}, sclera-to-eyelid boundaries were created using a linear fit between adjacent points marked on the eyelids. It is perhaps for this reason that the use of nearest-neighbor interpolation outperformed bilinear interpolation in the process of upsampling. Although the smoother curves that result from bilinear interpolation resulted in more accurate detection of the iris and pupil, it was less accurate in segmentation of the sclera.

Finally, data prepossessing had a significant impact on model performance. Introduction of CLAHE and gamma correction resulted in an overall improvement of 0.2\% in the validation mIoU score. Augmentation helped in noisy cases such as reflections from eyeglasses, varying contrast, eye makeup, and other image distortions.

\section{Conclusion}
We designed a computationally efficient model for the segmentation of eye images. We also presented methods for implementing multiple loss functions that can tackle class imbalance and ensures crisp semantic boundaries. We showed several methods for incorporating pre-processing and augmentation techniques that can help mitigate against image distortions. RITNet attained 95.3\% on the OpenEDS test set with a model size $<$1 MB and benchmarks an impressive 301Hz on a NVIDIA 1080Ti.

\section*{Acknowledgements}
We thank Anjali Jogeshwar, Kishan KC, Zhizhuo Yang, and Sanketh Moudgalya for providing valuable input and feedback. We would also like to thank the Research Computing group at RIT for providing access to GPU clusters.

{\small
\bibliographystyle{ieee}
\bibliography{Main}
}
\end{document}